\pdfoutput=1

\documentclass[11pt]{article}

\usepackage[]{EMNLP2022}

\usepackage{times}
\usepackage{latexsym}

\usepackage[T1]{fontenc}

\usepackage[utf8]{inputenc}

\usepackage{booktabs}
\usepackage{multirow}
\usepackage{xspace}
\usepackage{changepage}

\usepackage{amsfonts}
\usepackage{enumitem}

\usepackage{microtype}

\usepackage{inconsolata}

\usepackage[textsize=scriptsize]{todonotes}

\newcommand{\ben}{\textsc{AggreFact}\xspace}
\newcommand{\cnnben}{\textsc{AggreFact-Cnn}\xspace}
\newcommand{\cnnbensota}{\textsc{AggreFact-Cnn-FtSota}\xspace}
\newcommand{\xsumben}{\textsc{AggreFact-XSum}\xspace}
\newcommand{\xsumbensota}{\textsc{AggreFact-XSum-FtSota}\xspace}
\newcommand{\annotatedcnn}{\textsc{AggreFact-Cnn-Unified}\xspace}
\newcommand{\annotatedxsum}{\textsc{AggreFact-XSum-Unified}\xspace}
\newcommand{\cnnerror}{\textsc{AggreFact-Cnn-Error}\xspace}
\newcommand{\xsumerror}{\textsc{AggreFact-XSum-Error}\xspace}

\definecolor{blue}{HTML}{5c85a7}
\definecolor{red1}{HTML}{cc5e52}
\definecolor{purple}{HTML}{856b79}

\title{Understanding Factual Errors in Summarization: \\ Errors, Summarizers, Datasets, Error Detectors}

\author{Liyan Tang$^\diamondsuit$, Tanya Goyal$^\diamondsuit$, Alexander R. Fabbri$^\spadesuit$, Philippe Laban$^\spadesuit$, \\ \bf Jiacheng Xu$^{\diamondsuit, \spadesuit}$,  Semih Yavuz$^\spadesuit$, Wojciech Kryściński$^\spadesuit$, Justin F. Rousseau$^\diamondsuit$, Greg Durrett$^\diamondsuit$ \\
        $^\diamondsuit$The University of Texas at Austin  \quad $^\spadesuit$Salesforce AI Research\\   
\texttt{lytang@utexas.edu}
}

\begin{document}
\maketitle

\begin{abstract}
The propensity of abstractive summarization models to make factual errors has been studied extensively, including design of metrics to detect factual errors and annotation of errors in current systems' outputs. 
However, the ever-evolving nature of summarization systems, metrics, and annotated benchmarks makes factuality evaluation a moving target, and drawing clear comparisons among metrics has become increasingly difficult.
In this work, we aggregate factuality error annotations from nine existing datasets and stratify them according to the underlying summarization model. We compare performance of state-of-the-art factuality metrics, including recent ChatGPT-based metrics, on this stratified benchmark and show that their performance varies significantly across different types of summarization models. Critically, our analysis shows that much of the recent improvement in the factuality detection space has been on summaries from older (pre-Transformer) models instead of more relevant recent summarization models. We further perform a finer-grained analysis per error-type and find similar performance variance across error types for different factuality metrics. Our results show that no one metric is superior in all settings or for all error types, and we provide recommendations for best practices given these insights.\footnote{Code and data are available at \url{https://github.com/Liyan06/AggreFact}.}
 
%
%
%
%
\end{abstract}
\section{Introduction}\label{sec:introduction}
Although abstractive summarization systems \cite{liu-lapata-2019-text, lewis-etal-2020-bart, raffel2020exploring, pegasus} have improved dramatically in recent years, these models still often include factual errors in generated summaries \cite{kryscinski-etal-2020-evaluating, maynez-etal-2020-faithfulness}. 
A number of metrics have emerged to detect factuality errors, including methods based on sentence entailment \cite{kryscinski-etal-2020-evaluating}, finer-grained entailment \cite{goyal-durrett-2020-evaluating, zhao-etal-2020-reducing}, question generation and answering \cite{wang-etal-2020-asking, durmus-etal-2020-feqa, scialom-etal-2021-questeval}, and discrimination of synthetically-constructed error instances \cite{cao-wang-2021-cliff}. 
Despite recent analyses \cite{pagnoni-etal-2021-understanding, Laban2022SummaCRN}, reliably comparing these metrics remains difficult. 

\par
In this paper, we provide a new benchmark that allows for finer-grained comparison between different factuality systems. We aggregate 9 existing annotated factuality datasets to create our benchmark \textbf{\ben}. We stratify it according to the underlying summarization model, categorized into \textsc{FtSota}, \textsc{EXformer} and \textsc{Old} based on their development timeline (see Section~\ref{sec:benchmark}). First, we ask: \textbf{do factuality metrics perform equally well at  identifying errors from state-of-the-art summarization models and from earlier models?} For nine recent factuality metrics, including recent ChatGPT-based metrics, we show that metric performance varies substantially between different categories of summarization models. Most importantly, we found that the standard way of reporting improvements on category-agnostic benchmarks can be misleading, as most of these gains are on the \textsc{Old} or \textsc{EXformer} subset of the data which are less important to detect. On summaries generated by \textsc{FtSota} models, we found that there is no single metric that is superior in evaluating summaries from both the  CNN/DM \cite{hermann-2015-cnndm} and XSum \cite{narayan-etal-2018-dont} datasets.

\begin{table*}
\small
\centering
\begin{tabular}{@{}p{0.2\linewidth}p{0.18\linewidth}p{0.06\linewidth}p{0.05\linewidth}p{0.37\linewidth}@{}}
\toprule
\textbf{Dataset} & \textbf{Annotators} & \textbf{Kappa} & \textbf{Gran} & \multicolumn{1}{c}{\textbf{Annotation Scheme}} \\
\midrule
FactCC \newline \cite{kryscinski-etal-2020-evaluating} & 2 authors & - & summ & binary consistency label \newline (consistent/inconsistent)  \\
\midrule
Wang'20 \newline \cite{wang-etal-2020-asking} & 3 crowd-sourced annotators & 0.34/0.51 & sent & binary consistency label \newline (consistent/inconsistent) \\
\midrule
SummEval \newline \cite{fabbri-etal-2021-summeval} & 5 crowd-sourced annotators and 3 authors & 0.70 & summ & 5-point Likert scale \\
\midrule
Polytope \newline \cite{huang-etal-2020-achieved} & 3 trained annotators & - & span & \{addition, ommision, inaccuracy intrinsic, inaccuracy extrinsic, positive-negative aspect\} \\
\midrule
Cao'22 \newline \cite{cao-etal-2022-hallucinated} & 2 authors and 3 graduate students & 0.81 & entity & \{Non-hallucinated, Non-factual Hallucination, Intrinsic Hallucination, Factual Hallucination\} \\
\midrule
XSumFaith \newline \cite{maynez-etal-2020-faithfulness} & 3 trained annotators & 0.80 & span & \{intrinsic, extrinsic\} \\
\midrule
FRANK \newline \cite{pagnoni-etal-2021-understanding}& 3 crowd-sourced annotators & 0.53 & sent & \{RelE, EntE, CircE, OutE, GramE, LinkE, CorefE, OtherE, NoE\} \\
\midrule
Goyal'21 \newline \cite{goyal-durrett-2021-annotating} & 2 authors & - & span & \{intrinsic, extrinsic\} $\times$ \{entity, event, noun phrase, others\} \\
\midrule
CLIFF \newline \cite{cao-wang-2021-cliff} & 2 experts & 0.35/0.45 & word & \{intrinsic, extrinsic, world knowledge, correct\} \\
\bottomrule
\end{tabular}
\caption{Metadata of datasets in \ben. We report the annotator source, inter-annotator agreement, annotation granularity, and scheme for each dataset. Wang'20 and CLIFF reported kappa scores for XSum/CNNDM separately.}\label{tab:metadata}
\end{table*}

To better understand their behavior, we next analyze \textbf{what error types are different factuality metrics capable of identifying} (Section~\ref{sec:finegrainedanalysis}).
To do this, we leverage datasets from our benchmark that have fine-grained error annotations and unify these into a single taxonomy. 
We find that the error type distribution changes over time and even differs between annotations of the same summarization models across factuality datasets. 
Analysis of the factuality metrics shows that metrics claiming SOTA performance can identify each error type better in general, but all metrics differ significantly in how they perform on the same error types across CNN/DM and XSum.

We conclude with the following recommendations for best practices in this area:

\begin{enumerate}[leftmargin=*]

    \item \textbf{Evaluate factuality metrics on summaries generated by the state-of-the-art summarization models.} We found generally worse performance when evaluating factuality systems on summaries generated by \textsc{FtSota} models instead of less recent models (Section~\ref{sec:comparemetric}). We release \ben to support this, which combines existing benchmarks and stratifies them according to the base summarization model, summarization dataset and error types. We suggest future work to augment our benchmark with LLM-generated summaries, e.g. from ChatGPT, which is beyond the scope of this paper. 
    \item \textbf{Choose an appropriate factuality metric for your downstream task at hand.} No one metric is superior across all settings  (Section~\ref{sec:finegrainedanalysis}). Fine-grained insights offered by our benchmark can be useful to compare strengths of different factuality metrics and make this choice. 
    \item \textbf{Annotate error types consistently with prior work for better comparability.} We found that error type boundaries in existing works are not clear and are not easy to leverage for cross-dataset metric comparisons (Section~\ref{sec:finegrainedanalysis}). 
\end{enumerate}

We hope that our analysis can shed light on what comparisons practitioners should focus on, how to understand the pros and cons of different metrics, and where metrics should go next. Further, we hope that future work would extend this to incorporate diverse summarization domains such as dialogue summarization \cite{tang-etal-2022-confit, fabbri-etal-2021-convosumm, zhang-etal-2021-emailsum} and medical evidence summarization \cite{Tang2023}. These would have different error distributions, and annotated datasets are needed to perform a more comprehensive comparison and design domain-invariant factuality metrics.

\section{Benchmark}\label{sec:benchmark}
\subsection{Benchmark Standardization}
%
%
Current factuality metrics are evaluated without considering the types of summarization models used to generate the annotated summaries. 
In these annotated datasets, a large proportion of summaries are generated by older models, such as a pointer-generator network \cite{see-etal-2017-get}, that often make obvious errors that recent models do not make. 
\textbf{We hypothesize that current factuality systems primarily make progress in identifying factuality inconsistencies in summaries generated by out-of-date summarization models.} 
If this hypothesis is correct, comparing factuality systems on such datasets provide us less useful information on how these metrics perform on modern summarization systems.
\paragraph{Summarization datasets splits} We introduce a new benchmark \textbf{\ben} built on top of SummaC from \citet{Laban2022SummaCRN}. The benchmark \textbf{Aggre}gates nine publicly available datasets (see Table~\ref{tab:metadata}) that consist of human evaluations of \textbf{Fact}ual consistency on model generated summaries. We focus particularly on incorporating recent datasets annotated on top of state-of-the-art pre-trained Transformer models.

All datasets contain summaries generated from articles in CNN/DM and XSum. Given the unique characteristics of CNN/DM and XSum, our proposed benchmark includes two subsets, \cnnben and \xsumben, that evaluate the performance of factuality metrics on these two datasets separately (Table~\ref{ben_stats}; see also Table~\ref{tab:cnndm_stats} and \ref{tab:xsum_stats} in the Appendix). This facilitates a more fine-grained and rigorous analysis of the metric performance.
\par 
Our benchmark formulates factual consistency evaluation as a binary classification task, following \citet{Laban2022SummaCRN}. The binary factuality labels for the summaries are determined by human evaluations on the annotated datasets (Section~\ref{sec:annotateddatasets}).

\begin{table}[t!]
    \small
    \centering
    \begin{tabular}{c|cc|cc|cc}
    \toprule
     & \multicolumn{2}{c|}{\textsc{\textbf{Old}}} & \multicolumn{2}{c|}{\textsc{\textbf{EXformer}}} & \multicolumn{2}{c}{\textsc{\textbf{FtSota}}} \\
        & val & test  & val & test  & val & test \\ \midrule
        \textbf{-\textsc{Cnn}} & 2297 & 2166 & 275 & 375 & 459 & 559 \\
        \textbf{-\textsc{Xsum}} & 500 & 430 & 500 & 423 & 777 & 558 \\
    \bottomrule
    \end{tabular}
    \caption{Statistics of \cnnben and \xsumben. Details of individual annotated datasets can be found in Appendix Table~\ref{tab:cnndm_stats} and ~\ref{tab:xsum_stats}.} \label{ben_stats}
\end{table}

\paragraph{Summarization model splits} To validate our hypothesis and make a careful comparison of factuality metrics, we further divide models that were used to generated summaries in the benchmark into three distinct categories: $C=\{$ \textsc{FtSota}, \textsc{EXformer}, \textsc{Old} $\}$, as seen in Table~\ref{ben_stats}. \textsc{FtSota} represents state-of-the-art fine-tuned summarization models, including BART \cite{lewis-etal-2020-bart}, PEGASUS \cite{pegasus} and T5 \cite{raffel2020exploring}. 
\textsc{EXformer} is a collection of early Transformer-based summarization models. 
Typical models that fit into this category include BERTSum \citep{liu-lapata-2019-text}, and GPT-2 \citep{radford-2019-gpt2}. 
The remaining models, such as Pointer-Generator \cite{see-etal-2017-get} and BottomUp \cite{gehrmann-etal-2018-bottom}, are instances of \textsc{Old}. 
A full description of the models in each category is found in Appendix~\ref{sec:model_category}.
\subsection{Benchmark Datasets} \label{sec:annotateddatasets}
The \textsc{SummaC} benchmark \citep{Laban2022SummaCRN} includes six annotated datasets for factual consistency evaluation. We directly include XSumFaith \cite{maynez-etal-2020-faithfulness}, FactCC \cite{kryscinski-etal-2020-evaluating}, SummEval \cite{fabbri-etal-2021-summeval}, and FRANK \cite{pagnoni-etal-2021-understanding} from \textsc{SummaC} in our benchmark. We do not include the CoGenSumm \cite{falke-etal-2019-ranking} dataset as the original task is ranking pairs of generated summaries instead of detecting factually consistent summaries, and pairs of summaries can be both factually consistent or inconsistent. We modify the Polytope \cite{huang-etal-2020-achieved} dataset in \textsc{SummaC} where we view summaries annotated with \emph{addition}, \emph{omission} or \emph{duplication} errors as factually consistent since these three error types are not related to factual consistency. We use the validation and test splits from \textsc{SummaC} for the above mentioned datasets.

In addition to modifying \textsc{SummaC}, we further include four annotated datasets. For Wang'20 \cite{wang-etal-2020-asking}, CLIFF \cite{cao-wang-2021-cliff} and Goyal'21 \cite{goyal-durrett-2021-annotating}, we create data splits based on the parity of indices, following \textsc{SummaC}. For Cao'22 \cite{cao-etal-2022-hallucinated}, we use the existing splits from the original work.

\paragraph{Deduplication and label disagreement correction} Some examples may be labeled for errors in multiple datasets. We removed all duplicates so that each instance appears only once in our benchmark. During this deduplication process, we detected 100 instances of the same summaries that are annotated in different datasets with \emph{different} factual consistency labels. 98 of them are between FRANK and XSumFaith, and 2 of them are between FRANK and SummEval. The authors of this work manually corrected the labels for these examples based on our judgment. 
\subsection{Benchmark Evaluation Metrics}

We use balanced accuracy to evaluate the performance of factuality metrics due to the imbalance of factually consistent and inconsistent summaries. 
We refer readers to \citet{Laban2022SummaCRN} for further justification of balanced accuracy as the evaluation metric. 
In each dataset, a factuality metric selects a threshold for \textsc{FtSota}, \textsc{EXformer} and \textsc{Old}, respectively, based on the performance on the corresponding validation set. 
The chosen thresholds convert raw scores from metrics into binary labels for balanced accuracy evaluation. 
We provide a weighted average of performance across all datasets in the benchmark (see Table~\ref{tab:cnndm_xsum_eval}). 
\section{Comparison of Factuality Metrics}\label{sec:comparemetric}
First, we evaluate several SOTA factual consistency metrics on our benchmark, namely \textbf{DAE} \cite{goyal-durrett-2020-evaluating, goyal-durrett-2021-annotating}, \textbf{QuestEval} \cite{scialom-etal-2021-questeval}, \textbf{SummaC-ZS}, \textbf{SummaC-Conv} \cite{Laban2022SummaCRN} and \textbf{QAFactEval} \cite{Fabbri2021}.\footnote{We do not consider other common metrics like ROUGE \cite{lin-2004-rouge}, BLEU \cite{papineni-etal-2002-bleu} or BERTScore \cite{bert-score} as prior work \cite{Fabbri2021} has shown that they have low correlation with factuality.} We also benchmark recent ChatGPT-based evaluation metics from \citet{luo2023chatgpt} and \citet{wang2023chatgpt}. \textbf{ChatGPT-ZS} and  \textbf{ChatGPT-CoT} \cite{luo2023chatgpt} prompt LLMs to directly output a binary factuality decision. On the other hand, \textbf{ChatGPT-DA} and \textbf{ChatGPT-Star} \cite{wang2023chatgpt} ask LLMs to score the factuality of generated summaries on a scale of 0-100 and 1-5 respectively. More details about these metrics, including exact prompts are included in Appendix~\ref{sec:metrics}.

\paragraph{Unifying these metrics} 
We consider each metric as a function $f(d,s) \rightarrow y$, mapping each (document, summary) pair to a score $y \in \mathbb{R}$. We convert each method into a binary classifier $f'(d,s) \rightarrow \{0,1\}$ by picking a threshold $t$ such that we predict 1 if $f(d,s) > t$ and 0 otherwise.\footnote{\textsc{ChatGPT-ZS} and \textsc{ChatGPT-CoT} do not require thresholds as they directly predict factual consistency labels.}

All thresholds are set separately for each metric. We consider two ways of setting the threshold for a metric: \textbf{threshold-per-dataset} and \textbf{single-threshold}. 
The first setting has thresholds $\{t^{m}_{d, c}\}$ within each metric for every dataset we consider, where $d, c$ and $m$ are any dataset in $D$, any model category from $C$, and any factuality metric, respectively. This allows one to choose the right metric for the task at hand.
The \textbf{single-threshold} setting defines one threshold $\{t^{m}\}$ per metric.

\begin{figure}
    \centering
    \includegraphics[width=0.9\linewidth, trim=120mm 0mm 180mm 0mm,clip]{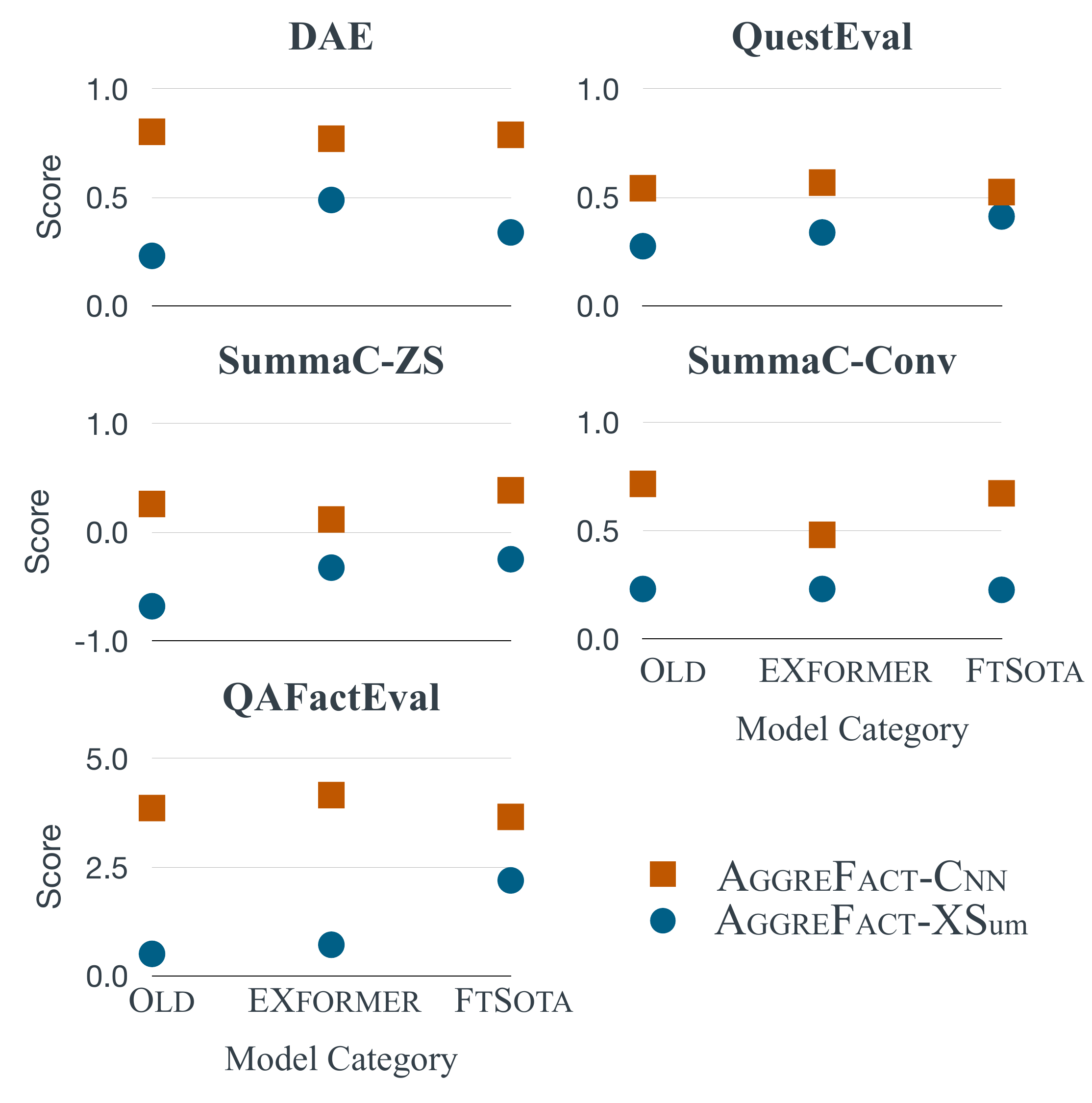}
    \caption{Average threshold values on \cnnben and \xsumben.}
    \label{fig:threshold}
\end{figure}

\paragraph{Threshold Analysis} \label{para:threshold} We analyze scores from factuality metrics using chosen thresholds $\{t^{m}_{d, c}\}$ from the validation sets. 
Specifically, for each factuality metric, we average the values of thresholds for each of \textsc{Sota}, \textsc{EXformer} and \textsc{Old} across all datasets (Figure~\ref{fig:threshold}). 
For all facuality metrics, the average threshold values for \cnnben are greater than those for \xsumben across all categories.
\textbf{This discrepancy of threshold values shows that evaluating on both of these datasets with a single threshold is a difficult balancing act and may lead to poor results on at least one dataset.}
\par
The higher threshold values on CNN/DM are connected to both the nature of the errors involved and overall extractiveness of the summaries 
XSum summaries are more abstractive and tend to contain a larger number of errors, making it harder for the metrics to verify the consistency of summaries with respect to the source text and resulting in lower scores in general, even for factual cases. For CNN/DM, smaller deviations from the source may indicate non-factuality.
\begin{table}
\centering
\renewcommand{\tabcolsep}{0.8mm}
\small
\begin{tabular}{lccc|ccc}
\toprule
 & \multicolumn{3}{c}{\textbf{\cnnben}} & \multicolumn{3}{c}{\textbf{\xsumben}} \\
 \cmidrule(r){2-4} \cmidrule(r){5-7}
 & \textsc{FtSota} & \textsc{EXf} & \textsc{Old}  & \textsc{FtSota} & \textsc{EXf} & \textsc{Old} \\
\midrule
Baseline & 50.0 & 50.0 & 50.0 & 50.0 & 50.0 & 50.0 \\
\midrule
DAE$^*$ & 59.4 & 67.9 & \underline{\textbf{69.7}} & 73.1 & - & - \\
QuestEval & 63.7 & 64.3 & \underline{\textbf{65.2}} & 61.6 & 60.1 & 59.7 \\
SummaC-ZS & 63.3 & \textbf{76.5} & 76.3 & \underline{\textbf{56.1}} & 51.4 & 53.3 \\
SummaC-Cv & 70.3 & 69.8 & \underline{\textbf{78.9}} & 67.0 & 64.6 & \textbf{67.5} \\
QAFactEval & 61.6 & 69.1 & \underline{\textbf{80.3}} & \underline{\textbf{65.9}} & 59.6 & 60.5 \\
\midrule
ChatGPT-ZS & 66.2 & 64.5 & \underline{\textbf{74.3}} & 62.6 & \underline{\textbf{69.2}} & 60.1 \\
ChatGPT-CoT & 49.7 & 60.4 & \underline{\textbf{66.7}} & 56.0 & \underline{\textbf{60.9}} & 50.1 \\
ChatGPT-DA & 48.0 & 63.6 & \underline{\textbf{71.0}} & 53.6 & \underline{\textbf{65.6}} & 61.5 \\
ChatGPT-Star & 55.8 & 65.8 & \underline{\textbf{71.2}} & 57.7 & \underline{\textbf{70.6}} & 53.8 \\
\bottomrule
\end{tabular}
\caption{Balanced accuracy on \cnnben and \xsumben across factuality metrics (threshold-per-dataset setting). A trivial baseline that predicts all examples as factually (in)consistent reports a balanced accuracy of 50\%. Since DAE was trained on the human-annotated XSumFaith data \cite{goyal-durrett-2021-annotating} including \textsc{EXformer} (\textsc{EXf} in table) and \textsc{Old} summaries, we exclude those results for a fair comparison. Best performing metric across all three categories is highlighted in \textbf{bold}, and \underline{underlined} if it is significantly better than the second best metric according to a paired bootstrap test.} \label{tab:cnndm_xsum_eval}
\end{table}

\paragraph{Binary Classification Results}
A weighted average of performance in terms of balanced accuracy for \cnnben and \xsumben is shown in Table~\ref{tab:cnndm_xsum_eval}.\footnote{Dataset-wise comparison between factuality metrics is shown in Appendix Table~\ref{fig:datasetwise_scores}.} It shows results using both trained metrics (upper half) and ChatGPT-based metrics (bottom half).

\par
Our results show that for \cnnben, both trained and ChatGPT-based factuality metrics achieve the best performance in evaluating the summaries in \textsc{Old}. 
This result is intuitive: the summaries in \textsc{Old} contain obvious errors, such as repetition, that can be more easily detected compared to more nuanced errors made by more recent models. 
From Table~\ref{ben_stats}, the majority of annotated summaries are generated by models from \textsc{Old}, so category agnostic performance  evaluation will weight these more heavily.
\textbf{There is a significant performance drop when evaluating the CNN/DM summaries generated by models from \textsc{EXformer} or \textsc{FtSota} instead.} 
Approximately a 10\% balanced accuracy decrease on average occurs from \textsc{Old} to \textsc{FtSota}. 
Evaluating on entire datasets, as is standard in prior work, gives us limited information of how these metrics perform on the \textsc{FtSota} summaries that are of more interest.

We observe more mixed results for \xsumben. Here, the trained and ChatGPT-based metrics perform best on \textsc{FtSota} and \textsc{EXformer} respectively. In fact, the ChatGPT-ZS and ChatGPT-Star metrics report new state-of-the-art results for the \textsc{EXformer} category.\footnote{We found that using different prompts can substantially vary the performance of ChatGPT metrics on both datasets. In our work, we use the exact same prompts as the original papers. Check the prompts in Appendix~\ref{sec:metrics}.} In the case of \xsumben also, we advocate for comparing metrics according to such a category-wise view as it provides more information on the most suitable metric to use while evaluating a given category of models.

%
%
%

\begin{table}
\small
\centering
\begin{tabular}{lcc}
\toprule
 & \begin{tabular}[c]{@{}c@{}}\textbf{\textsc{AggreFact-}}\\\textbf{\textsc{Cnn-FtSota}}\end{tabular} & \begin{tabular}[c]{@{}c@{}}\textbf{\textsc{AggreFact-}}\\\textbf{\textsc{XSum-FtSota}}\end{tabular} \\
\midrule
DAE & 65.4 $\pm$ 4.4 & \textbf{70.2 $\pm$ 2.3} \\
QuestEval & \textbf{70.2 $\pm$ 3.2} & 59.5 $\pm$ 2.7 \\
SummaC-ZS & 64.0 $\pm$ 3.8 & 56.4 $\pm$ 1.2 \\
SummaC-Conv & 61.0 $\pm$ 3.9 & 65.0 $\pm$ 2.2 \\
QAFactEval & 67.8 $\pm$ 4.1 & 63.9 $\pm$ 2.4 \\ 
\midrule
ChatGPT-ZS & 56.3 $\pm$ 2.9 & 62.7 $\pm$ 1.7 \\
ChatGPT-COT & 52.5 $\pm$ 3.3 & 55.9 $\pm$ 2.1 \\
ChatGPT-DA & 53.7 $\pm$ 3.5 & 54.9 $\pm$ 1.9 \\
ChatGPT-Star & 56.3 $\pm$ 3.1 & 57.8 $\pm$ 0.2\\
\bottomrule
\end{tabular}
\caption{Balanced binary accuracy using a single threshold on the \textsc{FtSota} subset (single-threshold setting). We show 95\% confidence intervals. Highest performance is highlighted in \textbf{bold}.} \label{tab:sota_eval}
\end{table}

\paragraph{Binary Classification: \textsc{FtSota}} To encourage comparison of factuality metrics on \textsc{FtSota} summaries, we provide a separate benchmark which consists of two subsets \cnnbensota and \xsumbensota that only consider summaries generated by \textsc{FtSota} models. This benchmark consists of validation and test splits from the \textsc{FtSota} subsets of the two datasets. This setting allows for comparisons of metrics to be made using only a single threshold. 
\par 
We show metric comparisons on the \textsc{FtSota} subset in Table~\ref{tab:sota_eval}. 
Note that the ranking of factuality metric here (single-threshold setting) is slightly different from the ranking in Table~\ref{tab:cnndm_xsum_eval}  (threshold-per-dataset setting). 
For \cnnbensota, QuestEval achieves the best performance amongst all metrics. We did not observe a statistically significant improvement over other trained evaluation metrics; however, its improvement over ChatGPT-based metrics is statistically significant. For \xsumbensota, the DAE metric is significantly better than all other metrics. 

Interestingly, metrics such as SummaC-Conv, QAFactEval and the recent ChatGPT metrics were all proposed as improved factuality evaluation on the category-agnostic SummaC benchmark (different from the SummaC metric). However, our stratified analysis provides a much clearer picture and shows that \textbf{metrics which claim improved performance on \textsc{SummaC} do not show similar gains when evaluated on \textsc{FtSota} summaries.} We recommend that future work similarly focuses on the \textsc{Sota} category of generated summaries when comparing factuality metrics. 

\section{Finer-grained Error Analysis}\label{sec:finegrainedanalysis}
Having established differences among factuality metrics across underlying summarization models, we now explore differences in metrics according to factuality error types.
To do this, we need a way to unify error types across datasets in our benchmark and map them into a shared taxonomy.

\begin{figure}[t]
\centering
    \includegraphics[width=\linewidth, trim=0mm 200mm 350mm 0mm,clip]{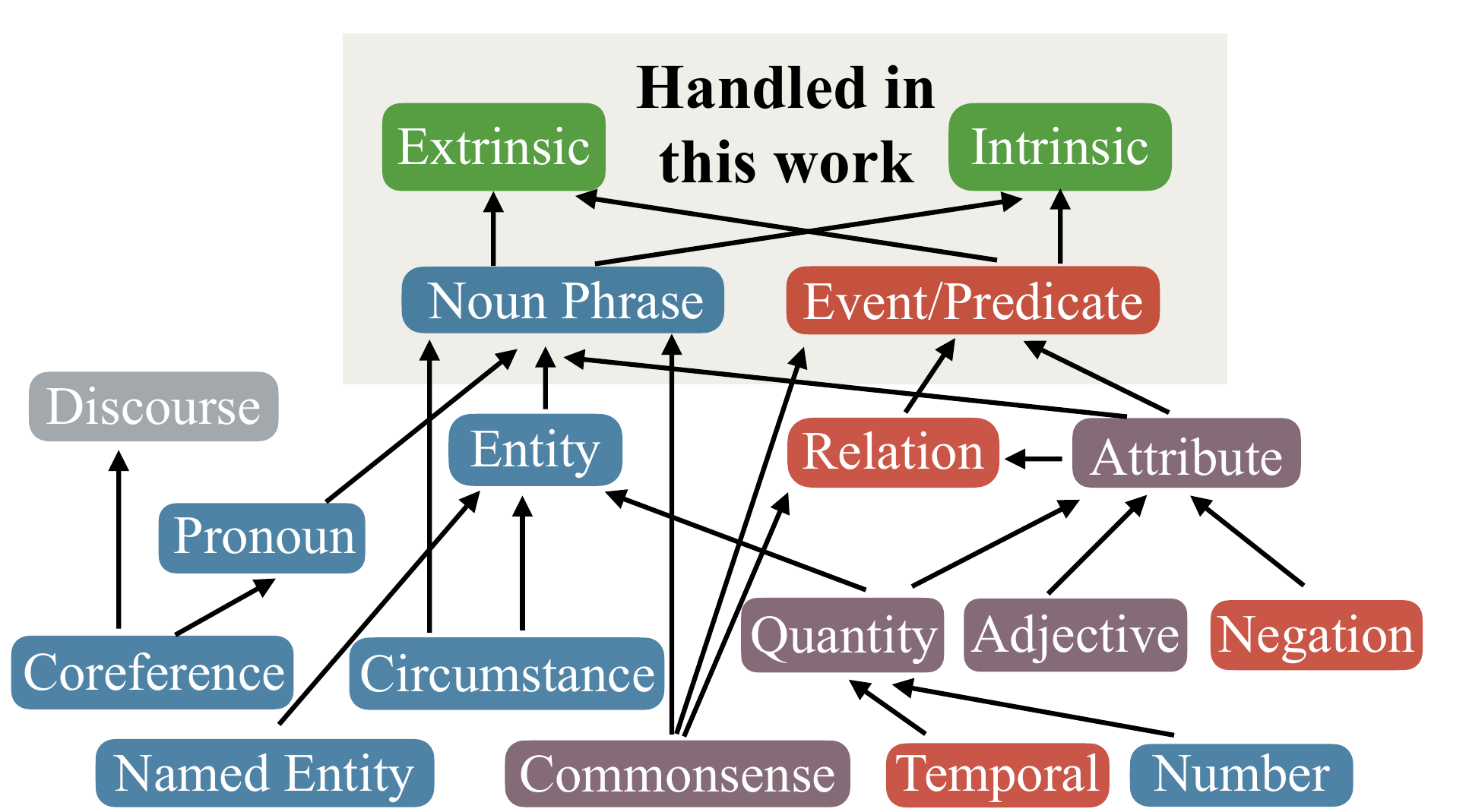}
    \caption{Taxonomy of factual consistency errors. We use unique colors to represent \setlength{\fboxsep}{1.2pt}\colorbox{blue}{\color{white}{entity}}- and  \setlength{\fboxsep}{1.2pt}\colorbox{red1}{\color{white}{predicate}}-related errors, as well as the \setlength{\fboxsep}{1.2pt}\colorbox{purple}{\color{white}{mix of two}}. See Appendix~\ref{sec:surveyerrortype} for citations of papers that use each error type.}
    \label{fig:error_taxonomy}
\end{figure}

\subsection{A Taxonomy of Error Types} \label{sec:types}
We surveyed existing error type taxonomies in prior work and unified the types of factual errors among them into a hierarchical taxonomy in Figure~\ref{fig:error_taxonomy}. 
Arrows relate more specific error types to more general ``parent'' errors.
The prior works that make use of each error type can be found in Appendix~\ref{sec:surveyerrortype}.
As shown in the figure, most error types related to factual consistency fall under the subset \emph{\{intrinsic, extrinsic\} $\times$ \{noun phrase, predicate\}} if we consider the coarsest level of the hierarchy. 
We discard discourse errors as these are uncommon and not present in most of our datasets.
Therefore, we consolidate all unique error type taxonomies from all four datasets we consider here into this error type subset (shown in the gray box in Figure~\ref{fig:error_taxonomy}). 
Descriptions and examples for these error types are in Table~\ref{tab:errordefinition}.
Further, we introduce two additional error categories \emph{\{intrinsic-entire sent., extrinsic-entire sent.\}} if an entire sentence is annotated as erroneous. 

\definecolor{Gray}{gray}{0.96}
\begin{figure*}
\centering
    \includegraphics[width=0.8\linewidth, trim=0mm 55mm 0mm 100mm,clip]{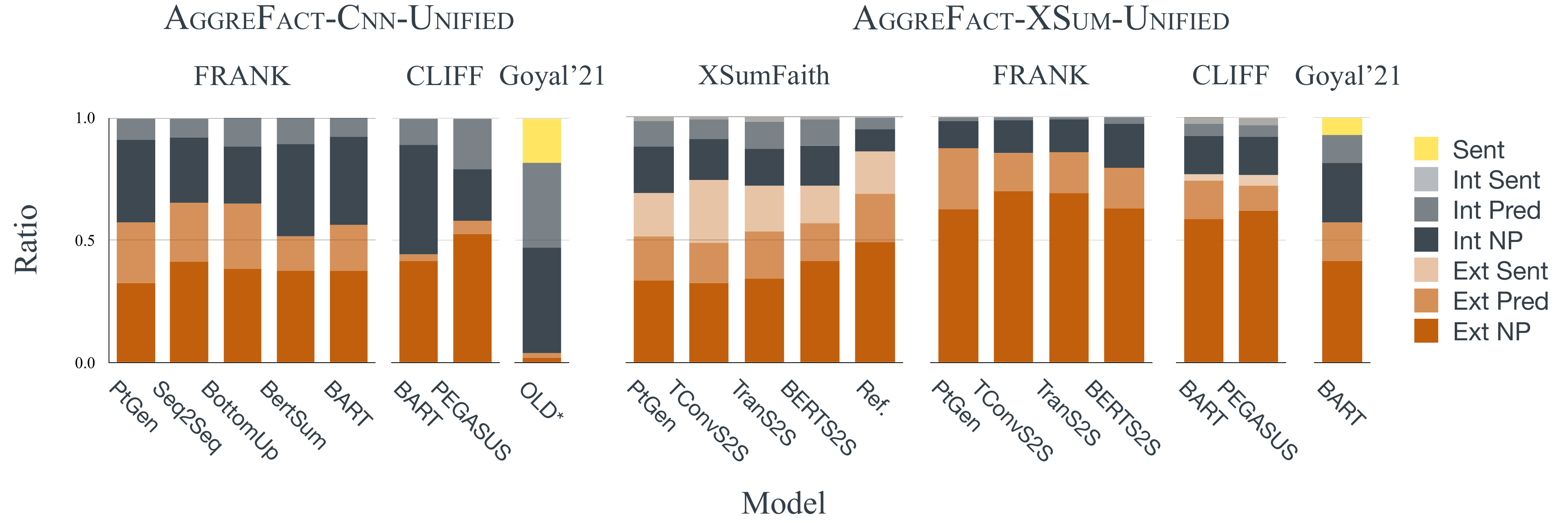}
    \caption{Error types of summaries from \annotatedcnn and \annotatedxsum. Ref. is annotated reference summary from XSumFaith. Since Goyal'21 in \annotatedcnn annotated summaries from FactCC, we use OLD$^*$ to denote summaries generated from \textsc{Old} models.}
    \label{fig:individualDistShift}
\end{figure*}

We are able to map four of the datasets (see Section \ref{sec:error-mapping}) in \ben that contain fine-grained annotations to our unified taxonomy.
For all four datasets, if there are multiple annotators, we assign an error type to a summary if the error is annotated by more than one annotator. We allow one summary to have multiple error types.
We call the annotated subset related to CNN/DM and XSum as \annotatedcnn and \annotatedxsum, respectively.

\subsection{Error Mapping}
\label{sec:error-mapping}
\paragraph{XSumFaith}
XSumFaith consists of 500 summaries each from human reference, two models in \textsc{Old}, and two models in \textsc{EXformer}. 
All summaries are annotated with intrinsic and extrinsic errors, but no finer categories are distinguished. 
For error type mapping, we automatically detect predicates in a summary and assign each error span intrinsic- or extrinsic-predicate error if it contains a predicate. 
We map the remaining error spans to intrinsic- or extrinsic-noun phrase error.
\paragraph{FRANK} 
The CNN/DM subset of FRANK consists of three models in \textsc{Old}, and one model each in both \textsc{EXformer} and \textsc{FtSota}. 
The XSum portion of FRANK has two models each in \textsc{Old} and \textsc{EXformer}. 
Each model contains 250 summaries in the dataset. 
We mapped Entity error and Out of Article error to extrinsic-noun phrase error; Predicate error and Grammatical error to extrinsic-predicate error; Circumstance error and Coreference error to intrinsic-noun phrase error; and other errors to intrinsic-predicate error.
\paragraph{Goyal'21} Authors of the original dataset manually identified all hallucinated text spans for each summary and classified hallucination types into \{intrinsic, extrinsic\} $\times$ \{entity, event, noun phrase, others\}.
The dataset consists of summaries for both CNN/DM and XSum. 
For the CNN/DM susbset, the authors directly annotated 50 summaries from FactCC, where summaries were generated by \textsc{Old} models. 
The XSum subset consists of summaries from \textsc{FtSota} models.
We map entity-related and noun phrase-related errors to noun phrase errors, event errors to predicate errors and others to entire sentence errors.
\paragraph{CLIFF} This dataset consists of 150 summaries each for both CNN/DM and XSum from two models in \textsc{FtSota}. 
We use the same approach for error mapping as we do for XSumFaith by only considering words labeled as extrinsic or intrinsic errors.

We evaluate the accuracy of our error type mapping via manual inspection. Specifically, the authors of this work inspect 30 factually inconsistent examples each for XSumFaith, FRANK and CLIFF. Those examples cover summaries generated by all models used in the datasets. Results of the manual inspection show that the accuracy of our error type mapping is over 90\%.
\par
A common discrepancy noticed by annotators was that in several cases the examples were originally annotated as intrinsic/extrinsic but we believe those errors are extrinsic/intrinsic. 
These cases are not a result of error in our mapping, but instead disagreement or error in the original annotation itself.
For error mapping, we found out mapping of FRANK to be least accurate among all 4 datasets.  For example, we found that the entity error (EntE) can be either intrinsic or extrinsic even though FRANK explicitly defines an extrinsic error type, i.e. ``out of article'' error. 
For Goyal'21, we manually correct any mapping errors that occur in the 150 examples.
Corrections mostly happen for the event-related error defined in Goyal'21 which can be either noun phrase- or predicate-related.

\subsection{Distribution Shift of Error Types} \label{sec:errorshift}
Next, we explore how the number of errors in specific groups of models from \textsc{FtSota}, \textsc{EXformer}, and \textsc{Old} has changed with the progress in the field. 
Specifically, for each of the FRANK, XSumFaith, Goyal'21, and CLIFF datasets, we calculate the ratio of error types from factually inconsistent summaries generated by each model. 
We then study any distribution shift of error types in \annotatedcnn and \annotatedxsum under \textsc{FtSota}, \textsc{EXformer}, and \textsc{Old}.

\paragraph{Summaries generated by the same models consist of different error distributions over different annotated datasets.} As shown in \annotatedxsum (Figure~\ref{fig:individualDistShift}), BART summaries are annotated by both Goyal'21 and CLIFF. 
However, it is interesting that BART summaries were annotated as making more intrinsic-noun phrase and intrinsic-predicate errors in Goyal'21 but more extrinsic-noun phrase errors in CLIFF. 
Similar observations can be found in \annotatedcnn, where BART summaries have a higher proportion of extrinsic-predicate error in FRANK and more intrinsic-noun phrase error in CLIFF.

\begin{table*}
\small
\centering
\begin{tabular}{lcccccccccc}
\toprule
 & \multicolumn{4}{c}{\textbf{\cnnerror}} & \multicolumn{6}{c}{\textbf{\xsumerror}} \\ 
 \cmidrule(r){2-5} \cmidrule(r){6-11}
 & \multicolumn{2}{c}{Intrinsic} & \multicolumn{2}{c}{Extrinsic} & \multicolumn{3}{c}{Intrinsic} & \multicolumn{3}{c}{Extrinsic} \\
  \cmidrule(r){2-3} \cmidrule(r){4-5} \cmidrule(r){6-8} \cmidrule(r){9-11}
 & \begin{tabular}[c]{@{}c@{}}NP\\ (183)\end{tabular} & \begin{tabular}[c]{@{}c@{}}Pred.\\ (60)\end{tabular} & \begin{tabular}[c]{@{}c@{}}NP\\ (220)\end{tabular} & \begin{tabular}[c]{@{}c@{}}Pred.\\ (129)\end{tabular} & \begin{tabular}[c]{@{}c@{}}NP\\ (196)\end{tabular} & \begin{tabular}[c]{@{}c@{}}Pred.\\ (113)\end{tabular} & \begin{tabular}[c]{@{}c@{}}Sent\\ (17)\end{tabular} & \begin{tabular}[c]{@{}c@{}}NP\\ (434)\end{tabular} & \begin{tabular}[c]{@{}c@{}}Pred.\\ (181)\end{tabular} & \begin{tabular}[c]{@{}c@{}}Sent\\ (197)\end{tabular} \\
\midrule
DAE$^*$ & 59.6 & 53.3 & 67.7 & 62.8 & - & - & - & - & - & - \\
QuestEval & 62.8 & 50.0 & 72.3 & 68.2 & 33.2 & 44.2 & 64.7 & 40.6 & 50.3 & 69.0 \\
SummacZS & 66.1 & \textbf{71.7} & \textbf{81.8} & 72.1 & 50.0 & 57.5 & 76.5 & 48.6 & 47.5 & 36.0 \\
SummacConv & 62.8 & \textbf{65.0} & 76.4 & 59.7 & 54.1 & 62.8 & 29.4 & 64.5 & 60.8 & 70.6 \\
QAFactEval & 56.3 & 51.7 & \textbf{79.1} & 63.6 & 66.8 & 75.2 & \textbf{88.2} & 55.1 & 70.2 & 79.2 \\
\midrule
ChatGPT-ZS & 56.3 & 45.0 & 63.2 & 52.7 & \textbf{83.2} & 85.8 & \textbf{94.1} & \textbf{74.2} & 83.4 & \textbf{93.9} \\
ChatGPT-COT & 54.1 & 60.0 & 61.8 & 52.7 & \textbf{83.2} & \textbf{91.2} & \textbf{94.1} & \textbf{77.2} & \textbf{89.5} & \textbf{91.9} \\
ChatGPT-DA & 65.0 & \textbf{73.3} & 71.8 & 67.4 & 55.6 & 67.3 & \textbf{94.1} & 53.7 & 65.7 & 67.5 \\
ChatGPT-Star & 65.0 & 68.2 & 68.2 & 56.6 & 66.8 & 73.5 & \textbf{94.1} & 64.7 & 74.6 & 75.1 \\
\bottomrule
\end{tabular}
\caption{Recall of factually incorrect summaries that contain certain error types (number of such summaries shown in parenthesis). Binary labels are directly obtained from \cnnben and \xsumben. We obtain 95\% confidence intervals and numbers in \textbf{bold} indicates that models have significantly higher recall of identifing certain error types compared to the rest of of the metrics. Since DAE is trained with human annotated data from XSumFaith, we remove DAE for a fair comparison in XSum error types.} \label{tab:errordetection}
\end{table*}
\par
In addition, although XSumFaith and FRANK annotate the same set of model generated summaries in \annotatedxsum, the distribution of error types looks dramatically different. The main discrepancy lies in the proportion of extrinsic-noun phrase and intrinsic-predicate errors. There are two possible reasons for such discrepancy. 
First, FRANK does not have ``entire sent.'' errors based on our conversation of its annotation schema to the unified taxonomy (Section~\ref{sec:error-mapping}).
Second, and more important, it is not easy to map error types from FRANK directly to our unified error types in spite of our validation.
For example, the ``out of article error'' in FRANK is defined as an error where some statements in the summary do not show up in the source text. 
We found this error can be mapped to either an extrinsic-noun phrase error or extrinsic-predicate error. 
These observations indicate that \textbf{previous work disagrees about where the individual error class boundaries are, even when aligned with our taxonomy}.

\begin{figure}
\centering
    \includegraphics[width=0.8\linewidth, trim=70mm 15mm 60mm 0mm,clip]{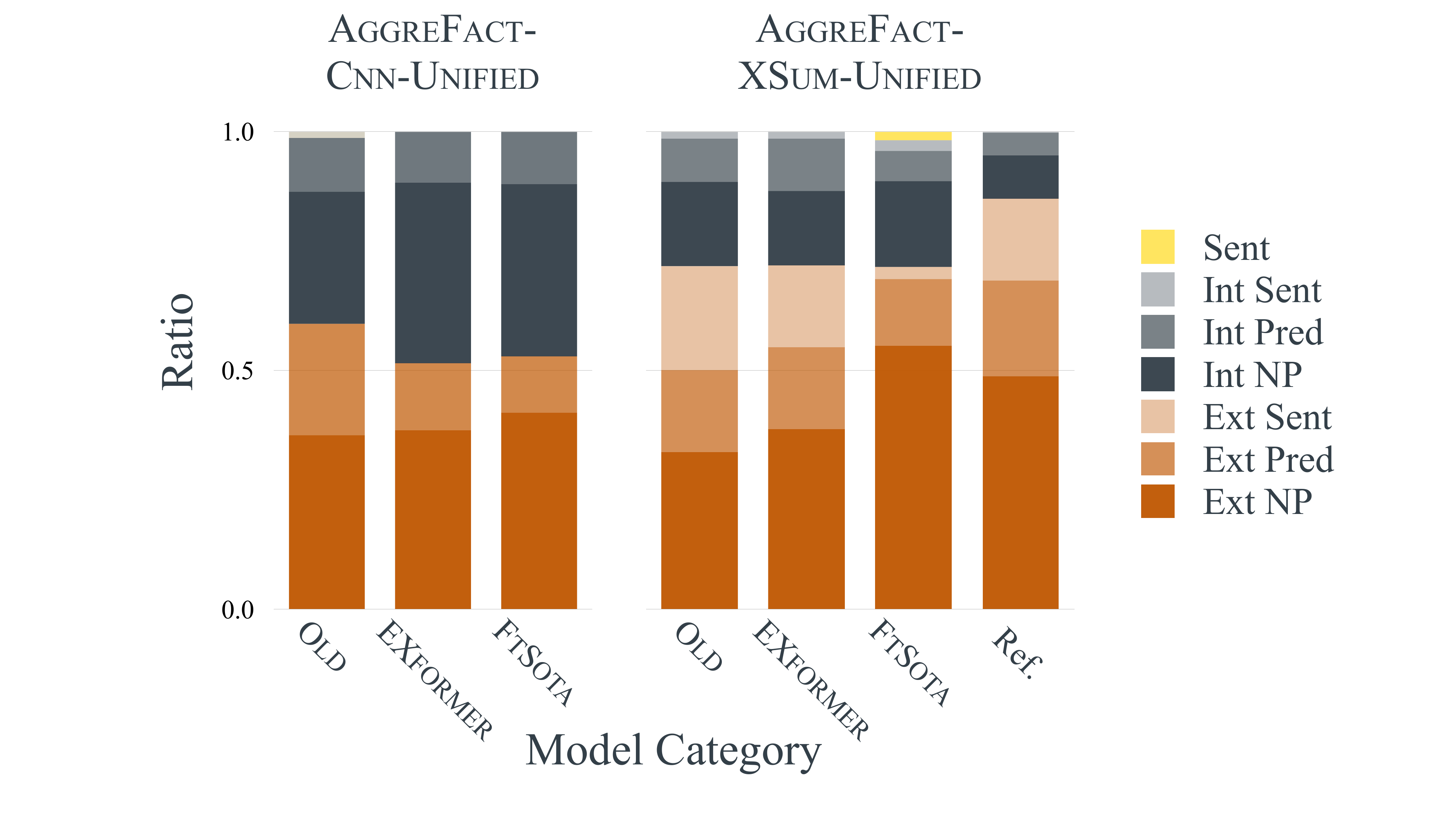}
    \caption{Distribution shift of error types on \annotatedcnn and \annotatedxsum. Ref. is human reference from XSumFaith.}
    \label{fig:distshift}
\end{figure}

\paragraph{A combined meta-analysis shows shifts in error distributions.} Figure~\ref{fig:individualDistShift} shows that error type distribution can vary among models from the same category. For example, summaries from BART contain a higher ratio of intrinsic-noun phrase errors than PEGASUS in \annotatedcnn.
We now combine all datasets together from \annotatedcnn and \annotatedxsum and show the unified error distributions over three model categories.\footnote{For \annotatedxsum, since XSumFaith and FRANK annotated the same set of summaries, we only use the annotation results from XSumFaith since our error mapping is more accurate on the span-level annotations.} 
As shown in Figure~\ref{fig:distshift}, models make approximately 50\% extrinsic errors in CNN/DM, with a slightly decrease from \textsc{Old} to more recent models. 
For XSum, the proportion of extrinsic errors remains unchanged and is at 70\%.

\subsection{Error Detection by Type}
In this section, we analyze how factuality metrics perform on summaries that contain certain error types. 
Specifically, we collect subsets of examples from four annotated datasets and group them into \cnnerror and \xsumerror.\footnote{We exclude FRANK for this analysis for the same reason as in Section~\ref{sec:errorshift}.} Every subset contains summaries that \textbf{include only one error type} defined in Section~\ref{sec:types}. 
Each factuality metric assigns a binary label to an instance obtained directly from \cnnben and \xsumben. 
Note that each subset only consists of test set examples from our benchmark since examples from the validation set were used to choose the optimal thresholds (Section~\ref{sec:comparemetric}). 
Since there are limited annotations for each model category after only considering examples from the test set of the benchmark, we decide not to split data by model categories in this part of the analysis. 
We calculate the recall of identifying error types from those subsets and show the results in Table~\ref{tab:errordetection}. 
Summaries in \cnnerror and \xsumerror primarily come from non-\textsc{FtSota} models (89.6\% and 92.1\%, respectively). 
On \cnnerror{}, where 79.0\% of summaries were generated from \textsc{Old}, there are more extrinsic errors (349) than intrinsic errors (243). 
This agrees with our above analysis that also shows that errors in generated summaries from less recent models are more likely to be extrinsic (Figure~\ref{fig:distshift}).

\par 
Across both \cnnerror and \xsumerror, we found that recent metrics like SummaC-Conv, QAFactEval and ChatGPT-based achieve higher recall for most error types.
This indicates that \textbf{more recent factuality metrics are better at capturing obvious errors generated by less recent models.}
This mirrors our earlier finding in Table~\ref{tab:cnndm_xsum_eval} (column \textsc{EXformer} and \textsc{Old}). 
Interestingly, we find that \textbf{summarization datasets (CNN/DM and XSum) have a non-negligible effect on the metrics' capabilities of detecting certain error types, even in the cases of out-of-date errors.} 
%
For example, the recall of identifying extrinsic-noun phrase error drops 10-30\% across all trained factuality metrics when evaluated on \textsc{XSum}, compared to \textsc{CNN/DM}. Similarly, ChatGPT metrics report 20-30\% higher recall on \textsc{CNN/DM}, compared to its \textsc{XSum} counterparts.

Another observation is that although DAE is trained using annotations from XSumFaith, which provides supervision for multiple error types, it does not identify errors as well in \cnnerror. 
These findings indicate that \textbf{summarization models make fundamentally different errors for each error type, and current factuality metrics cannot be uniformly good at identifying certain error types across datasets.} 
We believe this conclusion still holds when evaluating metrics on summaries generated from \textsc{FtSota} models since they generate less obvious errors.

\section{Recommendations}\label{sec:recommendations}
\paragraph{Evaluate factuality models on modern summarization systems} We have seen that \textsc{FtSota} yields significantly different results than \textsc{EXformer} or \textsc{Old}. 
Because of the prevalence of these systems, we believe that any new work should prefer evaluating on these \textsc{Sota} summaries. 

Particularly for factuality metrics that are either based on latest LLMs or on pre-trained models, evaluating on modern summarization systems is needed to see if these metrics are actually improving from the current state-of-the-art or merely patching errors in outdated systems that have already been fixed by other advances.

\paragraph{Annotate factual consistency errors from summaries generated by LLMs}
Recent work \cite{goyal2022news} shows that LLMs like GPT-3 are capable of generating summaries that are preferred over \textsc{FtSota} summaries by human annotators. Furthermore, they show that existing factuality metrics cannot reliably detect errors in summaries from GPT-3 models as these latter summaries differ substantially from existing benchmarks and training sets. We encourage future work to annotate errors from LLM-generated summaries and evaluate new factual consistency metrics on this set as well in addition to the \textsc{FtSota} set.  As such, we believe that future work should construct ``living'' benchmarks for factuality evaluation that are consistently updated as more powerful summarization systems are introduced.

\paragraph{Choose the right metric for the job} We note that there is no one clear winner among the metrics evaluated here (Section~\ref{sec:comparemetric}).
Depending on the downstream application, different methods may be more or less appropriate, as our analysis shows.
Moreover, none of current factuality metrics can identify certain error types across datasets equally well.
As QG/QA and NLI models get better, we expect all of these methods to improve further. Alternatively, although recent ChatGPT-based metrics \cite{luo2023chatgpt, wang2023chatgpt} do not perform well on modern summarization systems, they can be a starting point for leveraging LLMs to perform factual consistency evaluation.
\paragraph{Use more consistent error types} 
With our taxonomy, we have mapped error types annotated in previous work.
It is relatively easier and more accurate to map errors from XSumFaith, Goyal'21, and CLIFF to our unified error types as they have annotation granularity finer than sentence-level. 
We encourage future work to follow this taxonomy where possible and leverage definitions in prior work to make \emph{cross-dataset} comparisons possible.
Here also, we encourage future work to prioritize annotation and evaluation of SOTA summaries. 
\paragraph{Annotate and evaluate on non-news datasets} Most of current annotated datasets are within the news domain and factuality metrics are evaluated on news summaries accordingly.
As there is a rising interest in other domains such as dialogue summarization \cite{tang-etal-2022-confit, fabbri-etal-2021-convosumm, zhang-etal-2021-emailsum}, and medical evidence summarization \cite{Tang2023}, future work could annotate and analyze errors made by SOTA models there.
We encourage future work to develop factuality metrics that have superior performance over cross-domain evaluation.
\section{Conclusion}
In this work, we analyzed several factuality metrics across a large meta-benchmark assembled from existing datasets. 
We find state-of-the-art fine-tuned summarization models still present challenges for detecting factual errors, and the performance of error detectors is often overestimated due to the reliance on older datasets.
Furthermore, we unify existing datasets into a common taxonomy and use this to highlight differences between datasets and summarization models, as well as the complexity of unifying concepts in this problem space. 
\section*{Limitations}
There are a few limitations of our work. First, we focus on evaluating state-of-the-art factuality metrics on English newswire datasets. This setting restricts us to English-language data, a formal style of text, and topics consisting of what is discussed in US and UK-centric news sources. Moreover, other summarization domains such as dialogue summarization have different common error types such as \emph{wrong reference error} \cite{tang-etal-2022-confit}, which are not fully evaluated under current metrics. As settings like this are studied in future work, we believe that the kinds of analysis we do here can be extended to these settings as well.

Second, since our work is built on top of previous work, some analysis such as the error type mapping is limited by the quality and annotation agreement from previous work. We chose not to undertake large-scale reannotation to avoid causing confusion in the literature with multiple versions of datasets reflecting divergent annotator opinions. In spite of these limitations, we believe that our re-evaluation of these metrics and the analysis of error types under newswire data can bring insights for future works in choosing, designing and evaluating factuality metrics.
\section*{Acknowledgments}

The UT Austin team on this work was supported by a gift from Salesforce Inc., NSF Grant IIS-1814522, and a gift from Amazon.

\bibliography{anthology,extbib}
\bibliographystyle{acl_natbib}

\appendix
\section{Model Categories} \label{sec:model_category}

In this section, we briefly describe the summarization models we use in this paper. 

For \textsc{FtSota}, we include Transformer-based pre-trained models like BART \cite{lewis-etal-2020-bart}, T5 \cite{raffel2020exploring}, and PEGASUS \cite{pegasus}. They are pre-trained on massive text corpus and further fine-tuned on summarization datasets.


For \textsc{EXformer}, we use BERTSumExt and BERTSumAbs from \citet{liu-lapata-2019-text}, GPT-2 \cite{radford-2019-gpt2}, TransS2S \cite{vaswani2017attention}, and BERTS2S \cite{devlin-etal-2019-bert}.


For \textsc{Old}, we include models FastAbsRl \cite{chen-bansal-2018-fast}, TConvS2S \cite{narayan-etal-2018-dont}, BottomUp \cite{gehrmann-etal-2018-bottom}, PGNet \cite{see-etal-2017-get}, NeuSUM \cite{zhou-etal-2018-neural-document}, BanditSum \cite{dong-etal-2018-banditsum}, SummaRuNNer \cite{Nallapati}, TextRank \cite{mihalcea-tarau-2004-textrank}, CBDec \cite{jiang-bansal-2018-closed}, RNES \cite{rnes}, ROUGESal \cite{pasunuru-bansal-2018-multi}, ImproveAbs \cite{kryscinski-etal-2018-improving}, MultiTask \cite{guo-etal-2018-soft}, and UnifiedExtAbs \cite{hsu-etal-2018-unified}.

\section{Factuality Metrics} \label{sec:metrics}

We show the descriptions of consistency metrics we considered in our benchmark.

\paragraph{DAE} \cite{goyal-durrett-2020-evaluating} propose an arc entailment approach that evaluates the factuality $F_a(a,x) = P(\mathrm{entailment} \mid a, x)$ of each dependency arc $a \in \mathrm{Arc}(s)$ of the generated summary $s$ independently with respect to the input article $x$. It then uses their aggregation $\frac{1}{|\mathrm{Arc}(s)|} \sum_{a \in \mathrm{Arc}(s)} F_a(a,x)$ as the overall score. 
We use the default model and hyperparameters provided by the authors,\footnote{\url{https://github.com/tagoyal/factuality-datasets}} described in \citet{goyal-durrett-2021-annotating}, which is trained on data from XSumFaith, which we account for later in our comparisons. 
\paragraph{QuestEval} \cite{scialom-etal-2021-questeval} propose a QA-based metric that aggregates answer overlap scores from selected spans $r$ and questions $q_i \in Q_G(x)$ that derived from the input article $x$ and answered $Q_A(s, q_i)$ using the summary $s$ (recall-based); and those derived from the summary $q_i \in Q_G(s)$ and answered $Q_A(x, q_i)$ using the input article $x$ (precision-based). $Q_G$ and $Q_A$ denote question generation and question answering components, respectively. 
%
We use the implementation provided by the authors\footnote{\url{https://github.com/ThomasScialom/QuestEval}} and apply the unweighted version of the metric as in \citet{Laban2022SummaCRN}.
\paragraph{SummaC-ZS} \cite{Laban2022SummaCRN} is a zero-shot entailment metric that computes a sentence-level entailment score $F(s_i, x_j)$ between each summary sentence $s_i$ and input sentence $x_j$ using an NLI model $F$. It first find the maximum entailment score $\mathrm{score}(s_i) = \max_j F(s_i, x_j)$ for each summary sentence $s_i$, and averaging over all summary sentences for the final score $ \frac{1}{|s|}\sum_i \mathrm{score}(s_i)$. We use the default model and hyperparameters provided by the authors, which may return a negative score.
 
\paragraph{SummaC-Conv} \cite{Laban2022SummaCRN} extends SummaC-ZS by replacing the max operation with a binning of the entailment scores between each summary sentence $s_i$ and all input sentences $x_j$ to create a histogram $\mathrm{hist}(s_i, x)$. The histogram is then passed through a learned 1-D convolution layer $\mathrm{Conv}$ to produce the summary sentence score $\mathrm{score}(s_i) = \mathrm{Conv}(\mathrm{hist}(s_i, x))$. Parameters for the convolution layer are learned on synthetic data from FactCC \cite{kryscinski-etal-2020-evaluating}.
\paragraph{QAFactEval} \cite{Fabbri2021} is a QA-based metric analogous to the precision-based component of QuestEval and includes optimized question answering, generation, and answer-overlap components.
We do not make use of the variation of QAFactEval which combines QA and entailment-based scores into a single metric. 

\paragraph{ChatGPT-ZS} \cite{luo2023chatgpt} uses a zero-shot template and directly asks for a binary label of summary factuality.
\begin{adjustwidth}{1em}{1em}
\quad \emph{Decide if the following summary is consistent with the corresponding article. Note that consistency means all information in the summary is supported by the article.}

\emph{Article: [Article]}

\emph{Summary: [Summary]}

\emph{Answer (yes or no):}
\end{adjustwidth}

\paragraph{ChatGPT-CoT} \cite{luo2023chatgpt} also uses a zero-shot template but invokes chain-of-thought (CoT) style reasoning in its prompt. Similar to ChatGPT-ZS, it directly asks for a binary factuality label for a given summary.
\begin{adjustwidth}{1em}{1em}
\quad \emph{Decide if the following summary is consistent with the corresponding article. Note that consistency means all information in the summary is supported by the article.}

\emph{Article: [Article]}

\emph{Summary: [Summary]}

\emph{Explain your reasoning step by step
then answer (yes or no) the question:}
\end{adjustwidth}

\paragraph{ChatGPT-DA} \cite{wang2023chatgpt} uses a direct assessment (DA) prompt template that asks to assign a factual consistency score to a summary on a continuous scale from 0 to 100.
\begin{adjustwidth}{1em}{1em}
\quad \emph{Score the following news summarization given the corresponding news with respect to consistency on a continuous scale from 0 to 100, where a score of zero means ``inconsistency'' and score of one hundred means ``perfect consistency''. Note that consistency measures whether the facts in the summary are consistent with the facts in the original article. Consider whether the summary does reproduce all facts accurately and does not make up untrue information.}

\emph{Article: [Article]}

\emph{Summary: [Summary]}

\emph{Scores:}
\end{adjustwidth}

\paragraph{ChatGPT-Star} \cite{wang2023chatgpt} is an alternative version of ChatGPT-DA that asks LLMs to score summaries on a scale of one-to-five.
\begin{adjustwidth}{1em}{1em}
\quad \emph{Score the following news summarization given the corresponding news with respect to consistency with one to five stars, where one star means ``inconsistency'' and five stars means ``perfect consistency''. Note that consistency measures whether the facts in the summary are consistent with the facts in the original article. Consider whether the summary does reproduce all facts accurately and does not make up untrue information.}

\emph{Article: [Article]}

\emph{Summary: [Summary]}

\emph{Stars:}
\end{adjustwidth}

\section{Surveyed Error Types} \label{sec:surveyerrortype}

Here are our surveyed error types that are related to factual inconsistency.

\paragraph{Negation Error}	\cite{zhang-etal-2020-optimizing, kryscinski-etal-2020-evaluating, huang-etal-2020-achieved, zeng-etal-2021-gradient}
\paragraph{Adjective Error} \cite{zhang-etal-2020-optimizing}
\paragraph{Coreference Error} \cite{zhang-etal-2020-optimizing, kryscinski-etal-2020-evaluating, pagnoni-etal-2021-understanding, nan-etal-2021-improving}
\paragraph{Number error} \cite{kryscinski-etal-2020-evaluating, nan-etal-2021-improving, chen-etal-2021-improving, cao-etal-2020-factual}
\paragraph{Entity error} \cite{kryscinski-etal-2020-evaluating, pagnoni-etal-2021-understanding, zeng-etal-2021-gradient, wang-etal-2020-asking, nan-etal-2021-improving, nan-etal-2021-entity, chen-etal-2021-improving, cao-etal-2020-factual}
\paragraph{Attribute error}  \cite{pagnoni-etal-2021-understanding, huang-etal-2020-achieved}
\paragraph{Pronoun error} \cite{kryscinski-etal-2020-evaluating, zeng-etal-2021-gradient, cao-etal-2020-factual}
\paragraph{Commonsense error} \cite{kryscinski-etal-2020-evaluating}
\paragraph{Temporal error} \cite{kryscinski-etal-2020-evaluating, cao-etal-2020-factual}
\paragraph{Predicate error} \cite{pagnoni-etal-2021-understanding}
\paragraph{Discourse link Error} \cite{pagnoni-etal-2021-understanding}
\paragraph{Relation error} \cite{ nan-etal-2021-entity, nan-etal-2021-improving}
\paragraph{Quantity error} \cite{zhao-etal-2020-reducing}
\paragraph{Event error} \cite{goyal-durrett-2021-annotating}, 
\paragraph{Noun phrase error} \cite{wang-etal-2020-asking, goyal-durrett-2021-annotating}, 
\paragraph{Circumstance error} \cite{pagnoni-etal-2021-understanding}


%
\begin{table*}
\small
\centering
\begin{tabular}{lccccccccc}
\toprule
\multicolumn{2}{c}{} & Polytope & FactCC & SummEval & FRANK & Wang'20 & CLIFF & Goyal'21 & Total \\
\midrule
\multirow{2}{*}{\textsc{Old}} & val & 450 & 931 & 550 & 223 & 118 & - & 25 & 2297 \\
 & test & 450 & 503 & 548 & 523 & 117 & - & 25 & 2166 \\
\midrule
\multirow{2}{*}{\textsc{Xformer}} & val & 150 & - & 50 & 75 & - & - & - & 275 \\
 & test & 150 & - & 50 & 175 & - & - & - & 375 \\
\midrule
\multirow{2}{*}{\textsc{Sota}} & val & 34 & - & 200 & 75 & - & 150 & - & 459 \\
 & test & 34 & - & 200 & 175 & - & 150 & - & 559 \\
\bottomrule
\end{tabular}
\caption{Statistics of \cnnben. Each dataset is stratified into three categories \textsc{Old}, \textsc{EXformer}, and \textsc{FtSota}.} \label{tab:cnndm_stats}
\end{table*}
\begin{table*}
\small
\centering
\begin{tabular}{lccccccc}
\toprule
\multicolumn{2}{c}{} & XsumFaith & Wang'20 & CLIFF & Goyal'21 & Cao'22 & Total \\
 \midrule
\multirow{2}{*}{\textsc{Old}} & val & 500 & - & - & - & - & 500 \\
 & test & 430 & - & - & - & - & 430 \\
 \midrule
\multirow{2}{*}{\textsc{Xformer}} & val & 500 & - & - & - & - & 500 \\
 & test & 423 & - & - & - & - & 423 \\
 \midrule 
\multirow{2}{*}{\textsc{Sota}} & val & - & 120 & 150 & 50 & 457 & 777 \\
 & test & - & 119 & 150 & 50 & 239 & 558 \\
 \bottomrule
\end{tabular}
\caption{Statistics of \xsumben.} \label{tab:xsum_stats}
\end{table*}

\begin{table*}
\renewcommand{\tabcolsep}{1.1mm}
\small
\centering
\hspace*{-2.7em}
\begin{tabular}{clccccccccccc}
\toprule
\multicolumn{1}{l}{} &  & \multicolumn{1}{l}{} & \multicolumn{1}{l}{} & \multicolumn{9}{c}{\textbf{Factuality Metric}} \\ 
\cmidrule(r){5-13}
\multicolumn{1}{l}{} &  & \multicolumn{1}{l}{} & \multicolumn{1}{l}{} & \multicolumn{1}{l}{} & \multicolumn{1}{l}{} & \multicolumn{2}{c}{SummaC} & \multicolumn{1}{l}{} & \multicolumn{4}{c}{ChatGPT} \\ \cmidrule(r){7-8} \cmidrule(r){10-13}
\multicolumn{1}{l}{} &  & \multicolumn{1}{l}{} & \multicolumn{1}{l}{} & DAE & QuestEval & ZS & Conv & QAFactEval & ZS & COT & DA & Star \\
\multicolumn{1}{l}{} & \textbf{Dataset} & \textbf{Category} & \textbf{Count} &  &  &  &  &  &  &  &  &  \\
\midrule
\multirow{13}{*}{\begin{tabular}[c]{@{}c@{}}CNN\\ /DM\end{tabular}} & FactCC & \textsc{Old} & 503 & 0.704 & 0.655 & 0.835 & \textbf{0.891} & 0.843 & 0.793 & 0.697 & 0.686 & 0.743 \\ 
\cmidrule(r){2-13}
 & Wang'20 & \textsc{Old} & 117 & 0.586 & 0.552 & 0.655 & 0.672 & \textbf{0.754} & \textbf{0.758} & 0.599 & 0.695 & 0.652 \\ 
\cmidrule(r){2-13}
 & \multirow{3}{*}{SummEval} & \textsc{Old} & 548 & 0.661 & 0.649 & 0.773 & 0.801 & \textbf{0.814} & 0.735 & 0.680 & 0.735 & 0.713 \\
 &  & \textsc{EXformer} & 50 & 0.760 & 0.680 & 0.620 & 0.580 & 0.740 & 0.720 & 0.740 & \textbf{0.820} & 0.760 \\
 &  & \textsc{FtSota} & 200 & 0.452 & 0.649 & 0.622 & \textbf{0.827} & 0.652 & 0.783 & 0.401 & 0.453 & 0.568 \\ 
\cmidrule(r){2-13}
 & Polytope & \textsc{Old} & 450 & 0.779 & 0.687 & 0.802 & 0.791 & \textbf{0.824} & 0.768 & 0.695 & 0.741 & 0.752 \\
 &  & \textsc{EXformer} & 150 & 0.774 & 0.733 & \textbf{0.970} & 0.811 & 0.726 & 0.693 & 0.632 & 0.713 & 0.740 \\
 &  & \textsc{FtSota} & 34 & 0.294 & 0.176 & \textbf{0.971} & 0.735 & 0.324 & \textbf{0.941} & 0.735 & 0.206 & 0.412 \\ 
\cmidrule(r){2-13}
 & FRANK & \textsc{Old} & 523 & 0.704 & 0.669 & 0.692 & 0.728 & \textbf{0.773} & 0.694 & 0.628 & 0.695 & 0.672 \\
 &  & \textsc{EXformer} & 175 & 0.574 & 0.556 & 0.631 & 0.634 & \textbf{0.646} & 0.583 & 0.540 & 0.517 & 0.558 \\
 &  & \textsc{FtSota} & 175 & \textbf{0.699} & 0.626 & 0.570 & 0.601 & 0.547 & 0.519 & 0.514 & 0.523 & 0.531 \\
\cmidrule(r){2-13}
 & Goyal'21 & \textsc{Old} & 25 & 0.188 & 0.146 & 0.375 & 0.354 & 0.271 & 0.375 & 0.417 & \textbf{0.500} & \textbf{0.479} \\ 
\cmidrule(r){2-13}
 & CLIFF & \textsc{FtSota} & 150 & 0.730 & \textbf{0.740} & 0.646 & 0.649 & 0.716 & 0.603 & 0.550 & 0.528 & 0.612 \\
\midrule
\multirow{6}{*}{XSum} & Wang'20 & \textsc{FtSota} & 119 & \textbf{0.756} & 0.560 & 0.698 & 0.721 & \textbf{0.756} & 0.608 & 0.514 & 0.533 & 0.620 \\ 
\cmidrule(r){2-13}
 & Cao'22 & \textsc{FtSota} & 239 & \textbf{0.723} & 0.601 & 0.490 & 0.668 & 0.613 & 0.643 & 0.576 & 0.502 & 0.530 \\ 
\cmidrule(r){2-13}
 & XSumFaith & \textsc{Old} & 430 & - & 0.597 & 0.533 & \textbf{0.675} & 0.605 & 0.601 & 0.501 & 0.615 & 0.538 \\
 &  & \textsc{EXformer} & 423 & - & 0.601 & 0.514 & 0.646 & 0.596 & \textbf{0.692} & 0.609 & 0.656 & 0.706 \\ 
\cmidrule(r){2-13}
 & Goyal'21 & \textsc{FtSota} & 50 & 0.644 & \textbf{0.814} & 0.466 & 0.552 & 0.754 & 0.581 & 0.585 & 0.597 & 0.666 \\ 
\cmidrule(r){2-13}
 & CLIFF & \textsc{FtSota} & 150 & \textbf{0.754} & 0.619 & 0.596 & 0.668 & 0.613 & 0.643 & 0.576 & 0.502 & 0.530 \\ 
\bottomrule
\end{tabular}
\caption{Dataset-wise comparsion between factuality metrics. Since DAE is trained with human annotated data from XsumFaith, we remove DAE for a fair comparison. The best performance is highlighted in \textbf{bold} for each row.}
\label{fig:datasetwise_scores}
\end{table*}

\begin{table*}
\centering
\small
\begin{tabular}{@{}p{0.11\linewidth}p{0.45\linewidth}p{0.36\linewidth}@{}} \toprule

\textbf{Error Type} & \textbf{Definition} & \textbf{Example of Generated Summaries} \\
\midrule
Intrinsic-Noun Phrase & A model misrepresents word(s) from the source text that function(s) in a summary as subject, object, or prepositional object. & The world's first subsea power hub which uses a \textcolor{red}{lithium-based drive system} to generate electricity is being tested off the west coast of orkney. \\
\midrule

Intrinsic-Predicate & A model misrepresents word(s) from the source text that function(s) in a summary as the main content verb or content like adverbs that closely relate to the verb. & A conservative mp \textcolor{red}{has resigned} from his constituency as part of an investigation into a \# 10.25 m loan to a football club.\\
\midrule
Extrinsic-Noun Phrase & A model introduces word(s) not from the source text that function(s) in a summary as subject, object, or prepositional object but cannot be verified from the source. & Shale gas drilling in lancashire has been suspended after a magnitude-\textcolor{red}{7.5} earthquake struck. \\
\midrule
Extrinsic-Predicate & A model introduces word(s) not from the source text that function(s) in a summary as the main content verb or content like adverbs that closely relate to the verb, but which cannot be verified from the source. & Folate - also known as folic acid - \textcolor{red}{should be added} to flour in the uk, according to a new study.\\
\bottomrule
\end{tabular}
\caption{Definition and examples of unified error types. Factually inconsistent spans are highlighted in \textcolor{red}{red}.} \label{tab:errordefinition}
\end{table*}

\end{document}